\pdfoutput=1

\documentclass[11pt]{article}

\usepackage{acl}

\usepackage{times}
\usepackage{latexsym}

\usepackage{times}
\usepackage{soul}
\usepackage{natbib} 
\usepackage{url}
\usepackage{xcolor}
\usepackage{graphicx}
\usepackage{amsmath}
\usepackage{amsthm}
\usepackage{subcaption}
\usepackage{booktabs}
\usepackage{algorithm}
\usepackage{algorithm}
\usepackage{algpseudocode}
\usepackage{amsmath}
\usepackage{cleveref}
\usepackage{amssymb}

\algblock{Input}{EndInput}
\algnotext{EndInput}
\algblock{Output}{EndOutput}
\algnotext{EndOutput}

\Crefname{equation}{Eq.}{Eqns.}
\Crefname{figure}{Fig.}{Figs.}
\creflabelformat{equation}{(#2#1#3)}

\usepackage[T1]{fontenc}

\usepackage[utf8]{inputenc}

\usepackage{microtype}

\title{Impossible Triangle: What's Next for Pre-trained Language Models?}

\author{
Chenguang Zhu, Michael Zeng \\
Microsoft Cognitive Services Research Group\\
{
\{chezhu, nzeng\}@microsoft.com
}
}

\begin{document}

\maketitle
\begin{abstract}
Recent development of large-scale pre-trained language models (PLM) have significantly improved the capability of models in various NLP tasks, in terms of performance after task-specific fine-tuning and zero-shot / few-shot learning. However, many of such models come with a dauntingly huge size that few institutions can afford to pre-train, fine-tune or even deploy, while moderate-sized models usually lack strong generalized few-shot learning capabilities.
In this paper, we first elaborate the current obstacles of using PLM models in terms of the \textit{Impossible Triangle}: 1) moderate model size, 2) state-of-the-art few-shot learning capability, and 3) state-of-the-art fine-tuning capability. We argue that all existing PLM models lack one or more properties from the Impossible Triangle.
To remedy these missing properties of PLMs, various techniques have been proposed, such as knowledge distillation, data augmentation and prompt learning, which inevitably brings additional work to the application of PLMs in real scenarios.
We then offer insights into future research directions of PLMs to achieve the Impossible Triangle, and break down the task into several key phases.
\end{abstract}

\section{Background}

In recent years, large-scale pre-trained language models (PLM) have significantly improved the performance on various NLP tasks. Starting from BERT \citep{devlin2018bert} and GPT-2 \citep{gpt2}, the the paradigm of self-supervised pre-training followed by supervised fine-tuning has achieved great success, refreshing state-of-the-art results in many NLP areas such as semantic similarity \citep{jiang2019smart}, machine reading comprehension \citep{yang2019xlnet}, commonsense reasoning \citep{xu2021human}, and text summarization \citep{zhang2020pegasus}. Moreover, the moderate size\footnote{In this paper, we refer to models with fewer than 1B parameters as ``moderate-sized'', where a single GPU today can usually handle the model fine-tuning process.} of these PLMs allow widespread and fast model fine-tuning and adaptation.

However, in many real, especially novel, NLP scenarios, there are extremely limited labeled data for effective fine-tuning due to either budget or time constraint. This stimulates the development of zero-shot and few-shot NLP models. Starting from GPT-3 \citep{gpt3}, super large-scale PLMs (SL-PLM) have manifested superior performance on general NLP tasks when given only task descriptions and possibly some manual examples \citep{chinchilla,gopher,palm}. This capability was not previously observed in moderate-sized PLMs.
Nevertheless, the unprecedented scale of these SL-PLMs have to a large degree hindered their widespread application. One can hardly acquire enough computational resources even to load such models, let alone efficient deployment and possible fine-tuning. 

Thus, we argue that currently there have been no instance of a light-weight PLM that has excellent performance in both supervised learning and zero/few-shot scenarios for general NLP tasks. This has led to a lot of additional work in the employment of these PLMs in real scenarios, detailed in the next section.

\section{Impossible Triangle}
\begin{figure}[t]
\includegraphics[width=0.58\textwidth,trim=4cm 7.7cm 0cm 4.0cm]{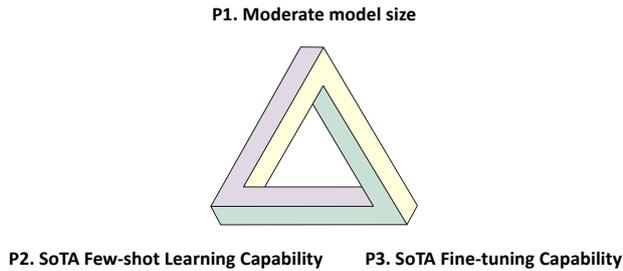}
\caption{The Impossible Triangle of pre-trained language models (PLM) consists of three desired properties for model deployment in real scenarios. \textbf{P1}: moderate model size, i.e., with fewer than 1 billion parameters, \textbf{P2}: state-of-the-art few-shot learning capability, and \textbf{P3}: state-of-the-art fine-tuning capability. (Triangle image is from \url{https://commons.wikimedia.org/wiki/File:Penrose_triangle.svg})}
\label{fig:triangle}
\end{figure}

In this section, we summarize the current obstacles of PLMs in the \textit{Impossible Triangle} (Fig. \ref{fig:triangle}). This triangle depicts the three key properties desired in PLMs for effective and efficient usage: \textbf{P1}: moderate model size, \textbf{P2}: state-of-the-art few-shot learning capability, and \textbf{P3}: state-of-the-art supervised learning capability. 
These three properties correspond to three requirements on the practical application of PLMs: 
P1 is for efficient deployment using a reasonable amount of computational resources; P2 is for scenarios with zero or very few labeled data; P3 is for scenarios with relatively abundant labeled data. 

One potential reason for the existence of the impossible triangle is that at the current stage strong few-shot learning power only emerges when a PLM has reached an immense size with enough model capacity. Although there are work such as iPET \citep{ipet} which designs moderate-sized PLMs to achieve better few-shot learning performances than GPT-3 \citep{gpt3}, they have been outperformed by later SL-PLMs such as PaLM \citep{palm}. And it is observed that there is discontinuous improvement of zero/few-shot performance with model scale \citep{palm}. For instance, compared to model variants with 8B and 62B parameters, PaLM with 540B parameters demonstrates a drastic jump in accuracy on many tasks. 
Thus, it remains a big challenge to develop a moderate-sized model with SoTA zero/few-shot learning performances while maintaining superb supervised learning capability.

Although no PLMs have achieved all three properties in the Impossible Triangle, many of them have acquired one or two of these capabilities:

\textbf{Moderate-size PLMs (With P1 + P3).} These language models have a moderate model size, i.e., fewer than 1 billion parameters, which enables efficient model tuning and deployment. They have set state-of-the-art results in general NLP tasks such as GLUE benchmark \citep{raffel2019exploring}, text summarization \citep{zhang2020pegasus}, open-domain question answering \citep{izacard2020distilling} and commonsense reasoning \citep{xu2021human}. Nevertheless, these models usually have relatively weak zero/few-shot capability, meaning that using such models depends on sufficient labeled data in the target domain.

\textbf{Super large-scale PLMs (With P2).} These language models have immense model size (1 to 1,000 billion parameters), and were pre-trained on very large-scale data. For instance, PaLM \citep{palm} with 540 billion parameters was pre-trained on a diverse text corpus with 780 billion tokens. They have achieved SoTA performance in general zero/few-shot NLP tasks, when prompted with only the task descriptions and possibly a few example input-output pairs. However, in general, i) the zero/few-shot performance of SL-PLMs is below that of supervised trained models, and ii) after fine-tuning, many SL-PLMs still underperform the best fine-tuned moderate-size PLMs \citep{hu2021lora}, possibly due to their immense model sizes\footnote{The most recent SL-PLM model PaLM has shown much better results on some NLP tasks such as summarization after fine-tuning.}.

\section{Current Remedies}
Because of the Impossible Triangle, many measures have been taken to address the missing capability(ies) of employed PLMs in practice. We summarize them as follows:

\textbf{Immense model size (Lack of P1).} This happens when a SL-PLM shows superb few-shot learning capability and also strong performance after fine-tuning.
To obtain a moderate-size model with a performance similar to that of the SL-PLM, a common practice is knowledge distillation (KD) \citep{gou2021knowledge}. In KD, the larger model acts a teacher and the smaller model is the student, learning from the predicted distribution and/or parameters from the teacher. Knowledge distillation has been effective in creating much more efficient models with a little sacrifice in performance. 

However, there remain two problems. First, the learned student can hardly attain the same performance of the teacher. Second, the immense size of SL-PLMs hinders efficient inference, making them inconvenient as teacher models.

\textbf{Inferior zero/few-shot performance (Lack of P2).} This is most common for moderate-sized PLMs which achieve SoTA performance after fine-tuning, but have relatively low zero/few-shot learning capability.
In many scenarios, one wants to deploy such models when there is a lack of sufficient labeled data. Thus, one way to remedy this problem is data augmentation \citep{feng2021survey}. By generating pseudo labels and pseudo data instances from other models \citep{wang2021want} or noise injection \citep{feng2021survey}, the model can leverage these additional data for effective supervised training. Nevertheless, the variation in pseudo data quality and the diversity of data types in different tasks pose a challenge to a generally applicable solution. 

\textbf{Inferior supervised training performance (Lack of P3).} This is typical when fine-tuning an SL-PLM, where the computational resource is limited or the amount of training data is insufficient for fine-tuning a super large model. A typical solution is prompt learning \citep{liu2021pre}. One can leverage either a hard prompt, i.e., discrete textual template, or a soft prompt, i.e., continuous parameter embeddings, so that only the words and in-context examples in hard prompt or parameters in the soft prompt are updated during fine-tuning \citep{liu2021makes,ptuning}. This has been shown to be quite effective to improve the performance of an SL-PLM given labeled data. However, the performance could be very sensitive to prompt selection and training data \citep{zhao2021calibrate} and be still inferior to that of a moderate-size PLM with supervised learning.

\section{Future}
While the Impossible Triangle currently exists for NLP models, we argue that it may be solved in a multi-phase approach.

\textbf{Phase 1.} PLMs are developed with the goal of achieving some desired properties in the triangle while improving on the other missing properties. For instance, a moderate-sized model with SoTA supervised learning capability could be improved on its few-shot learning performance; or a SL-PLM with SoTA few-shot learning capability is compressed into smaller models with better supervised learning performance.

\textbf{Phase 2.} PLMs achieving all three desired properties are developed for one of a few NLP tasks, such as NER or text summarization. To achieve this, one can leverage the unique characteristics of the target task, e.g., less dependency of the performance on the training data scale, smaller gap between zero/few-shot and supervised learning performances, etc. 

\textbf{Phase 3.} PLMs achieving all three desired properties on general NLP tasks are developed, built upon the progress in Phase 1 and Phase 2. Potential methods include pre-training a moderate-sized model with much larger data, better knowledge distillation, generalized data augmentation methods, etc.

Once a PLM is equipped with all three properties of the Impossible Triangle on general NLP tasks, it will transform the entire landscape of NLP research and applications, facilitating fast, efficient and high-quality model development and deployment.

\section{Conclusion}
To adapt to various practical scenarios, a pre-trained language model (PLM) needs to have a reasonable size and acquire superb zero/few-shot and supervised learning capabilities.
However, currently there exists no models that can achieve all these three desired properties, which we frame as the \textit{Impossible Triangle}.
To remedy the missing properties from Impossible Triangle, various technologies have been employed in practice, such as knowledge distillation, data augmentation and prompt learning.
We then propose future research directions of pre-trained language models to achieve the Impossible Triangle in three progressive phases.

\bibliography{acl}
\bibliographystyle{acl_natbib}

\end{document}